\documentclass[review]{elsarticle}

\usepackage{lineno,hyperref}
\usepackage{times}
\usepackage{epsfig}
\usepackage{graphicx}
\usepackage{amsmath}
\usepackage{amssymb}
\usepackage{multirow}
\usepackage{rotating}
\usepackage{booktabs}
\usepackage{xcolor}
\usepackage{float}
\usepackage{caption}
\usepackage{subcaption}
\usepackage{verbatim}
\newcommand{\norm}[1]{\left\lVert#1\right\rVert}
\usepackage{cleveref}
\crefname{section}{§}{§§}
\Crefname{section}{§}{§§}

\usepackage{amssymb}

\journal{Computer Vision and Image Understanding}

\begin{document}

\begin{frontmatter}



\title{Unsupervised Action Proposal Ranking through Proposal Recombination}

\author{ Waqas Sultani, Dong Zhang and Mubarak Shah}
 \address{Center for Research in Computer Vision, University of Central Florida, USA\\ waqassultani@knights.ucf.edu, dzhang@eecs.ucf,edu, shah@crcv.ucf.edu}

\author{}

\address{}

\begin{abstract}
Recently, action proposal methods have played an important role in action recognition tasks, as they reduce the search space dramatically. Most unsupervised action proposal methods tend to generate hundreds of action proposals which include many noisy, inconsistent, and unranked action proposals, while supervised action proposal methods take advantage of predefined object detectors (e.g., human detector) to refine and score the action proposals, but they require thousands of manual annotations to train. 

Given the action proposals in a video, the goal of the proposed work is to generate a few better action proposals that are ranked properly.  In our approach, we first divide action proposal into sub-proposal and then use Dynamic Programming based graph optimization scheme to select the optimal combinations of sub-proposals from different proposals and assign each new proposal  a score.  We propose a new unsupervised image-based  actioness detector that leverages web images and employ it as one of the node scores in our graph formulation. Moreover, we capture motion information by estimating the number of motion contours within each action proposal patch. The proposed method is an unsupervised method that neither needs bounding box annotations nor video level labels, which is desirable with the current explosion of large-scale action datasets. Our approach is generic and does not depend on a specific action proposal method. We evaluate our approach on several publicly available trimmed and un-trimmed datasets and obtain better performance compared to several proposal ranking methods. In addition, we demonstrate that properly ranked proposals produce significantly better action detection as compared to state-of-the-art proposal based methods.
\end{abstract}


\end{frontmatter}


\section{Introduction}

Spatio-Temporal detection of human actions in real-world videos is one of the most challenging problems in computer vision. Taking page from object detection in images,  several action detection methods have been introduced for videos \cite{Yicong-cvpr2013,Figurecentric2011,ActionTubes,Mihir,APT,Khurram}.  Following the success of sliding window-based approaches for images, authors in \cite{Yicong-cvpr2013,Figurecentric2011} used sliding windows for action detection/localization in videos. However, videos have much larger search space as compared to images (for instance, the number of 3D bounding boxes (or tubes) is $O(w^2h^2t^2)$ for a video of size $w \times h$ with $t$ frames). Moreover, these 3D boxes may contain large portion of background and cannot fully capture complex articulations of human actions. Therefore, several action proposal methods have been introduced \cite{Mihir,Danoneata,APT,FastProposals,FlatteningICCV2013} in recent years to reduce the search space and improve action localization accuracy. These methods rely heavily on video segmentation \cite{Danoneata,Mihir,FlatteningICCV2013}  or clustering of motion trajectories \cite{IIDTF,APT}. These methods not only improve efficiency but also improve the accuracy of detection by reducing false positive rate. Since these methods use hierarchical segmentation or clustering, they tend to generate thousands of action proposals in each video, where many of proposals are noisy and do not contain action. Moreover, these methods utilize low-level color, motion and saliency cues but ignore high-level action cues, which make them difficult to generate actioness score for each proposal. Hence, the classification methods treat all proposals equally.  Fully supervised action proposal methods such as \cite{FastProposals} employ thousands of manual human annotations and use motion information from training videos to obtain ranked action proposals. However, with the recent explosive growth of action datasets \cite{THUMOS14}, it is prohibitive to obtain bounding box annotations for each video, thus the application of fully supervised action proposal methods is limited.

In this paper, we address the above-mentioned limitations of action proposals methods. We do not propose any new action proposal method. Instead, given the output of any recent action proposal method  \cite{Danoneata,Mihir,APT},  our goal is to generate a few new action proposals which are properly ranked according to how well they localize an action. The proposed method assigns an accurate actioness score to each proposal, without using any labeled data (no human detection \cite{FastProposals} or bounding box annotations), thus is easier to generalize to other datasets.

The proposed approach begins with dividing each proposal into sub-proposals.  We assume that the quality of proposal remains the same within each sub-proposal. We, then employ a graph optimization method to recombine the sub-proposals in all action proposals in a single video in order to optimally build new action proposals and rank them by the combined node and edge scores. The node score is a combination of image-based actioness score (or image actioness; since this is computed for an image, not a video) and motion scores. The image-based actionnes scores are obtained by a deep network trained on web images. The training data is obtained without manual annotation by an automatic image co-localization method. In essence, the trained deep network is a generic image actioness detector rather than a specific action detector. The motion scores are obtained by measuring the number of motion contours enclosed by each proposal bounding box patch. The edge scores between sub-proposals are computed by the overlap ratio and appearance similarity of temporally neighboring sub-proposals. {Note that edges are made between temporally adjacent sub-proposals. For an untrimmed video, we first divide the video into shots and then make the above-mentioned graph within each shot}. Our method generates a few ranked proposals that can be better than all the existing underlying proposals. Our experimental results validated that the properly ranked action proposals can significantly boost action detection results. Our method is summarized in Figure 1.

The rest of the paper is organized as follows. In Section \ref{sec:related}, the related work on action and object proposals is introduced. Section \ref{sec:method} presents how we score and rank action proposals in order to obtain fewer but better proposals. In Section \ref{sec:exp}, we quantitatively evaluate proposal ranking, contributions of different components and effect of properly ranked proposals on action detection. We conclude in Section \ref{sec:conclusions}.

\begin{figure*}[t]
\begin{center}
\includegraphics[width=12cm]{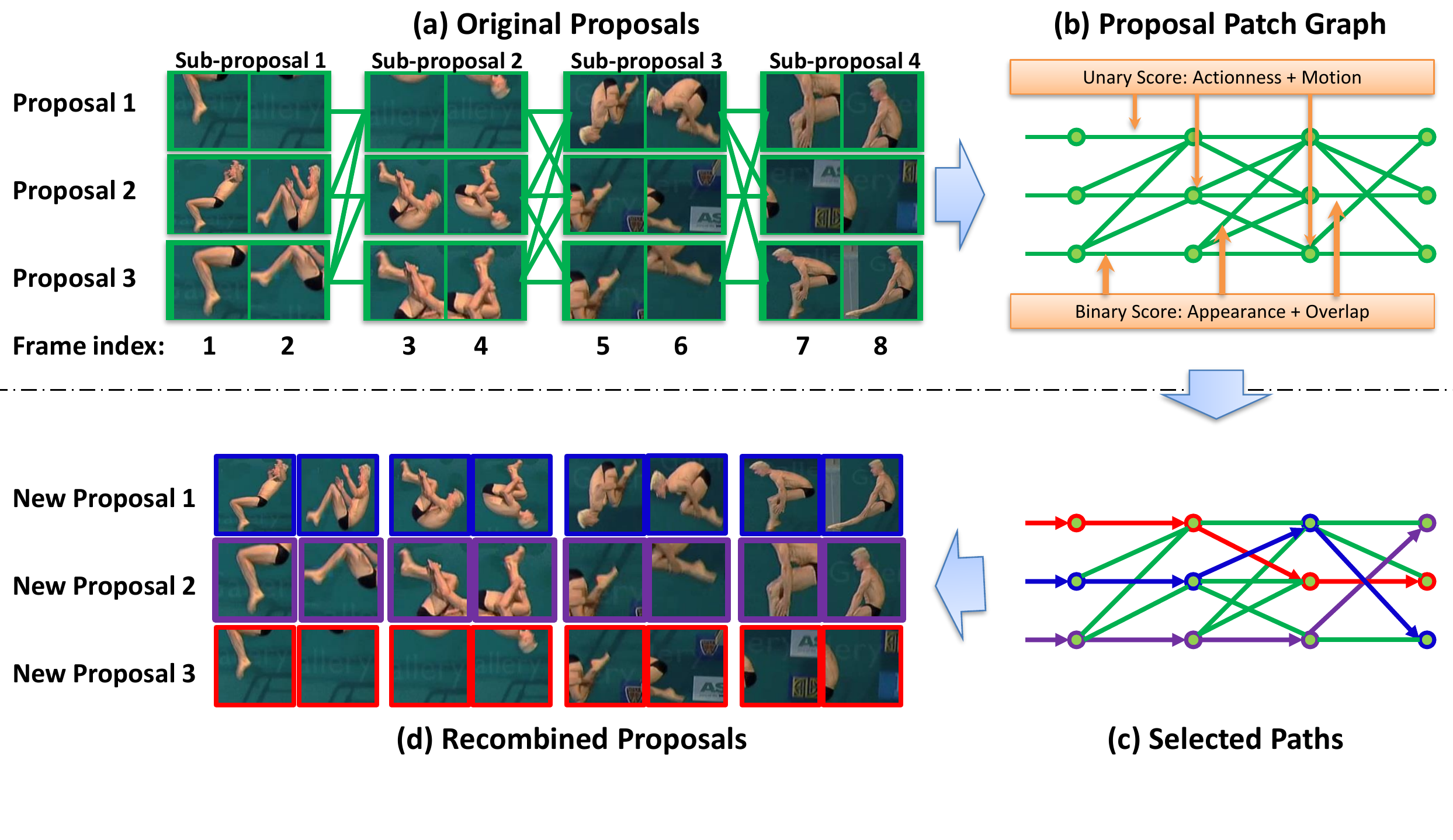}
\end{center}
   \caption{An illustration of the proposed method. In this illustration, there are 3 action proposals with 8 frames each. (a) shows the sub-proposal patches of the original action proposals. (b) shows the corresponding graph. Each node in (b) represents one sub-proposal, and edges represent the consistencies between the nodes. (c) shows the 3 top selected paths in the graph  (in the order of blue, red, and purple). (d) shows the ranked new action proposals corresponding to the graph in (c), and it is easy to see that these are much better than the original proposals in (a).}
\label{fig_framework}
\end{figure*}

\section{Related work}
\label{sec:related}

Till the last few years, most object detection methods used sliding window approaches \cite{DPM}. In order to reduce search space and computation complexity of sliding window approach, several improvements have been made \cite{Selective, Lampert}. The most popular approach in this direction is to use object proposals \cite{Objectness,ZitnickECCV14edgeBoxes, Selective, RandomPrim}. An object proposal provides category independent candidate object location which might contain an object. Due to their high recall and efficiency, it is not surprising that most of the successful object detection approaches use object proposals as their pre-processing steps \cite{RCNN}.

Although quite successful, the object proposal methods have two main limitations: 1) They tend to generate thousands of candidate object locations, where many of these locations either do not contain any object or enclose only a part of object; 2) Since proposals are used as a pre-processing step, if an object is missed at this step, irrespective of robustness of object detection method, that object can never be detected.  Ku et al. \cite{DeepBox} pointed out that the small number of properly ranked proposals can have a significant impact on efficiency and quality of object detection system. The authors in \cite{DeepBox} proposed to  rank object proposals using a  light weight CNN network trained on carefully obtained ground truth annotations. More recently, Zhong et al. \cite{ReRank} utilized semantic, contextual and low-level cues in a fully supervised approach to re-rank object proposals.


Following the line of object detection research, \cite{Yicong-cvpr2013,SubvolumeSearch} proposed to detect complex human actions in videos using sliding window search in spatio-temporal space. Tian et al. \cite{Yicong-cvpr2013} introduced 3D version of \cite{DPM} and used spatio-temporal parts to detect action.  Yu et al. \cite{SubvolumeSearch} used a  branch and bound method to efficiently search action volumes. As videos have much larger search space than images and fixed sized video sub-volumes are unable to capture spatio-temporal articulation of actions; several action proposal methods have been proposed recently\cite{Mihir,Danoneata,APT,FastProposals}. Following selective search based object proposal method \cite{Selective} in images, Jain et al. \cite{Mihir} presented a video based action proposal method using motion information. The method \cite{Mihir} starts with supervoxel segmentation \cite{GBH-ECCV2012}, followed by a hierarchical grouping method using motion, color and texture cues. The method shows improved action classification results compared to several sliding window alternatives. Similarly, Oneata et al. \cite{Danoneata} extend \cite{RandomPrim} from images to videos and introduced randomized supervoxel segmentation method for proposal generation. Although both methods \cite{Danoneata,Mihir} produce good quality proposals, they are computationally expensive due to their dependencies on supervoxel segmentation. Recently, \cite{APT,FastProposals} presented a faster method for generating action proposals by removing supervoxel segmentation altogether. Yu et al. \cite{FastProposals} generated action proposals using supervised human detector and connecting human detection boxes across multiple frames. In addition to action proposals, authors in \cite{FastProposals} also produce a probability of actioness for each proposal. Van Gemert et al. in \cite{APT} employ dense trajectory features \cite{IIDTF} and produce accurate action proposals by clustering \cite{SLINK} dense trajectory features. This method has shown good action classification results as compared to several baseline methods on many challenging datasets.  Inspired by the success of deep learning features for object detection \cite{RCNN}, Gkioxari et al. \cite{ActionTubes} proposed to use two stream network using object proposals \cite{Selective}. They link the high scored action boxes across frames to find action tubes. Weinzaepfel et al.  \cite{Weinzaepfel_2015_ICCV} employ tracking by detection approach to obtain action localization using CNN features and motion histograms. 

Similar to objectness score \cite{Objectness}, Wei et al. \cite{actionness} introduced a supervised method using lattice conditional random field to estimate the actioness of the region in frames of videos.  Limin et al. \cite{actionness16} presented hybrid convolution neural network based method to compute the probability of actioness. However, these methods require lots of carefully obtained manual annotations.

Similar to object proposals, action proposals tend to generate hundreds or thousands of action proposals; however, only a few of them contain action.  Moreover, while most object proposal methods \cite{Selective,DeepBox,ZitnickECCV14edgeBoxes,Objectness} produce objectness scores (or ranking) for the proposals; however, most action proposal methods \cite{Mihir,APT,Danoneata} do not produce actioness scores (or ranking), and \cite{FastProposals} is among the few outliers. In order to obtain a properly ranked action proposals and discover the most action representative proposal, Siva et al. \cite{NegativeMining} proposed to use negative mining approach. Specifically, they utilized videos with the labels other than the label of interest and select the proposal whose distance to the nearest neighbor in negative videos is the maximum. Tang et al \cite{TangCVPR13} presented an approach to obtain ranked proposals using negative videos similar to \cite{NegativeMining}, but they used more robust criterion by using all negative proposals. Using ranked proposals, they have shown excellent results for weakly supervised object segmentation in videos. Recently, Sultani and Shah  \cite{Waqas16_AL} used web images to rank action proposal in action videos. Although these methods have shown improved ranking results, they are weakly supervised and hence needs video level labels. Moreover, if the action location is missed by underlying proposal generation method, these ranking methods cannot discover missed action location as they just do the ranking.

In contrast, given action proposals, we propose to automatically obtain only a few properly ranked action proposals. Our approach is unsupervised  as it needs neither bounding box annotations nor video level labels. Moreover, due to the recombination of sub-proposals across different proposals, it has the ability to discover new proposals that can be better than all of the initially generated proposals.

\section{Action Proposal Recombination and Ranking}
\label{sec:method}

The proposed approach for action proposal ranking starts by obtaining candidate action proposals and then recombines them in order to get fewer but better-ranked proposals. Since the number of  candidate action proposals is huge in number and many proposals are noisy, in order to obtain robust ranking, we divide each proposal into sub-proposals and build a graph  across the proposals with the sub-proposals as the nodes. The node score of the graph is a combination of image-based actioness score and motion score. Edges between nodes impose the frame consistencies, and their scores are a combination of intersection-over-union and appearance similarity between sub-proposals. The action proposals are generated and ranked in an iterative manner: to maximize the node$+$edge scores.
The combined node+edge score is assigned to each proposal as its score. After selection of this proposal, all related nodes are removed from the graph and the next proposal is selected in the same way. This is an iterative process and it can produce and rank an arbitrary number of action proposals as needed. Note that our method is not limited to any specific methods for generating candidate action proposals; any recent action proposal method \cite{FastProposals,Danoneata,Mihir,APT} can be employed within the framework. In experiments, we demonstrate the superior performance of our approach using initial proposals obtained from three different proposal methods.

In what follows, we introduce the graph formulation in Section \ref{sec:graph}, and explain the node and edge scores, respectively, in Sections \ref{sec:node_score}, \ref{sec:motion_score}, and \ref{sec:edge_score}.

\subsection{Graph Formulation}
\label{sec:graph}

We formulate the problem of action proposal ranking as a graph optimization problem (Fig. \ref{fig_framework}). The graph $G=\{V,E\}$ consists of a set of nodes $V=\{v_i^f|i=1..N,f=1...F\}$ and edges $E=\{e_{i,j}^f|i,j=1...N,f=1...F\}$, where $i,j$ are the proposal indices, $f$ is the sub-proposal index, $N$ is the number of action proposals and $F$ is the number of sub-proposals in the video.  In order to obtain sub-proposals, we divide video into five equal temporal segments. { Note that the accuracy of our approach is stable across several different  number of sub-proposals.}

 Node scores are defined as
\begin{equation}
\label{equ_node_score}
    \Phi = \lambda^i \cdot \Phi^i + \lambda^m \cdot \Phi^m,
\end{equation}
where $\Phi^a$ and $\Phi^m$ are the image-based actioness score and motion score respectively. $\lambda^a$ and $\lambda^m$ are the corresponding weights for each term. The edge scores are defined as
\begin{equation}
\label{equ_edge_score}
    \Psi = \lambda^o \cdot \Psi^o + \lambda^a \cdot \Psi^a,
\end{equation}
where $\Psi^a$, $\Psi^o$ are the appearance similarity and shape consistency scores, and $\lambda^a$, $\lambda^o$ are the weight adjustments accordingly. Combining the node and edge scores we get as follows:
\begin{equation}
    \label{equ_graph_energy}
    E(P) = \sum_{f=1}^{F}(\Phi_{p_f}^f + \lambda \cdot \Psi_{(p_f,p_{f+1})}^f),
\end{equation}
where $P=\{p_1,...,p_F\}$ are the sub-proposal indices selected to make one proposals, $\Phi_i^f$ is the node score of $i$th proposal in $f$ sub-proposal, $\Psi_{(i,j)}^f$ is the edge score for the edge that connects two  $i$th and $j$th proposal in temporally adjacent sub-proposals, and $\Psi_{(.,.)}^F = 0$ (i.e. just to ensure the notation to be consistent for the last frame). The goal is:
\begin{equation}
    \label{equ_graph_solution}
    P^* = \arg \max_P E(P).
\end{equation}
Our aim now is to select optimal paths through this graph. The graph $G$ in Fig. \ref{fig_framework} is a directed acyclic graph and the solution for Eqn. \ref{equ_graph_solution} can be obtained in polynomial time using dynamic programming \cite{Yang2011}, we use the value of this function to rank different proposals. After each iteration, the selected nodes are removed, and the next subset of nodes is selected using the same process until a specific number of action proposals are obtained.

Each term of the node score $\Phi$ in Eqn. \ref{equ_node_score} and edge score $\Psi$ in Eqn. \ref{equ_edge_score} is introduced in next two sections.


\subsubsection{Image-based actioness score}
\label {sec:node_score}
The objective of image-based actioness score is to provide a probability score for each proposal patch to measure whether or not some action is being performed there. In contrast to generic object-ness, little work has been done for the generic image-based actioness score (also sometimes called actioness). It is well known that different actions share much high-level articulation and semantic information. Many actions, for instance, walking, jogging, running, kicking, diving, swinging, share similar patterns of motion in different images of the videos. Therefore, we believe that learning a generic image-based action detector for all actions can robustly provide us a probability of the presence of some action in an image.

Training a actioness detector from videos is a daunting task, as it requires a large number of spatio-temporal annotated frames of actions. In that case, generic action detector is practically useless since one would expect that training generic action detector should be easy as compared to training specific action detector. Fortunately, recent works \cite{Mihir_15,Waqas16_AL} have demonstrated that deep network trained on images can also provide good classification and localization results in videos. Since that obtaining the bounding box annotations of images is less expensive than for videos, we employ images for the actioness detector. It is still a cumbersome task to get enough bounding box image annotations for a deep network. Therefore, we leverage the internet to obtain relevant images. However, a simple search for ``human" and ``person" does not work well since it results in producing images contain faces or simple standing people. In contrast, by searching action terms (``man walking", ``cricket bowling", ``swing", etc.), top retrieved images contain good quality, well-centered action images that capture the key poses and distinct articulation of actions. Therefore, these images can be a good source for training an actioness detector.

\begin{figure*}
\begin{center}
\includegraphics[width=12cm]{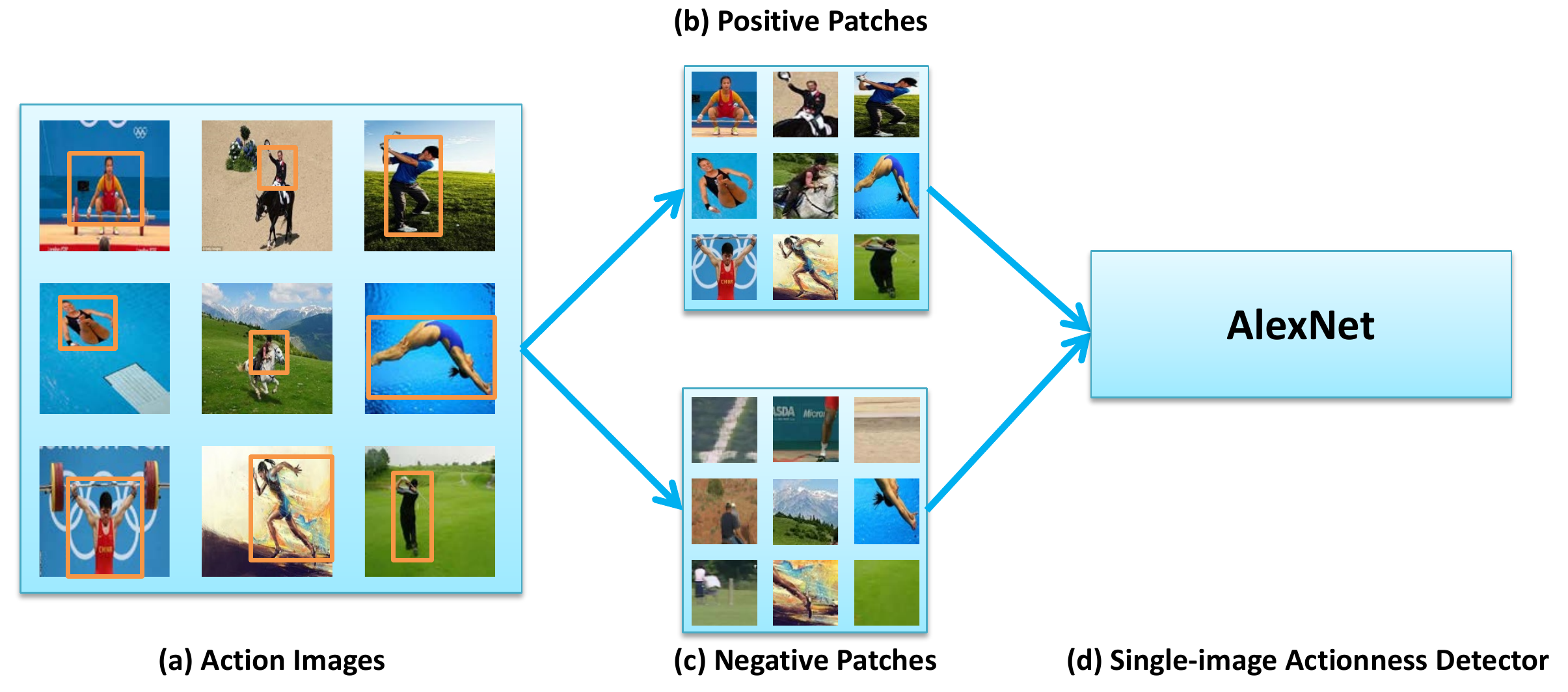}
\end{center}
   \caption{Single-image actioness detection. (a) shows some action images downloaded from Google. (b) shows the positive patches obtained by unsupervised bounding box generation. (c) shows some of the low optical flow negative patches obtained from UCF-Sports videos. (d) is the Alexnet based actioness detector.}
   \label{actionness}
\end{figure*}

\paragraph{Positive Images for Image Actioness Detector}
We use UCF-Sports (10 actions) \cite{action-mach-08} action names to download action images from Google. To avoid ambiguity in a text query, we changed some action names slightly, for example using, `soccer kicking' instead of `kicking', `men walking' instead of `walk'.  In order to cope with noisy images and backgrounds within images, we used following pre-processing steps \cite{Waqas16_AL}.

\paragraph{Noisy Image Removal}
Since most of the retrieved images belong to the same concept (the textual query), they form a dense cluster in a feature space \cite{NEIL}. On the other hand, outliers are usually far away from dense clusters or make small clusters. To remove these outliers, we employed random walk, which is well proven to be good at getting rid of outliers in such scenarios \cite{OulierDetection}. We represent each image with 16-layer VGG features, $\psi$,  \cite{vedaldi15matconvnet} and make a fully connected graph between all the images. The transition probability of random walk on this graph from image $i$ to image $j$ is given by
\begin{equation}
p(i,j)=\frac{e^{-\alpha \norm {\psi(i)-\psi(j)}_2}}{\sum_{m=1}^ke^{-\alpha \norm{\psi(i)-\psi(m)}_2}}.
\end{equation}
The random walk over the graph is then formulated as:
\begin{equation}
    s_k(j)= \gamma\sum_i s_{k-1}(i)p_{ij}+(1-\beta)z_j,
\end{equation}
where $s_k$ represents similarity score of image $j$ at $k^{th}$ iteration. We use the same initial probability $z_j$ for each image. After a fixed number of iterations, we remove the images that receive lower confidence.

\paragraph{Unsupervised Bounding Boxes Generation}

Action images that contain actors with significant background may hurt the performance of the classifier. In order to get rid of the background and capture actual action (the common pattern across images), we employ an unsupervised localization approach similar to \cite{Minsucho}. First, for all images of the same action, the nearest neighbors are computed using GIST features and then part based matching is performed by employing Probabilistic Hough Matching (PHM) in HOG feature space.
The final localization is the patches in the images that contain common pattern across images (action in action images). In order to further remove noisy patches, we repeat random walk framework (described above) on image patches and remove low consistency noisy patches.

\begin{figure*}
\begin{center}
\includegraphics[width=10cm,height=2.8cm]{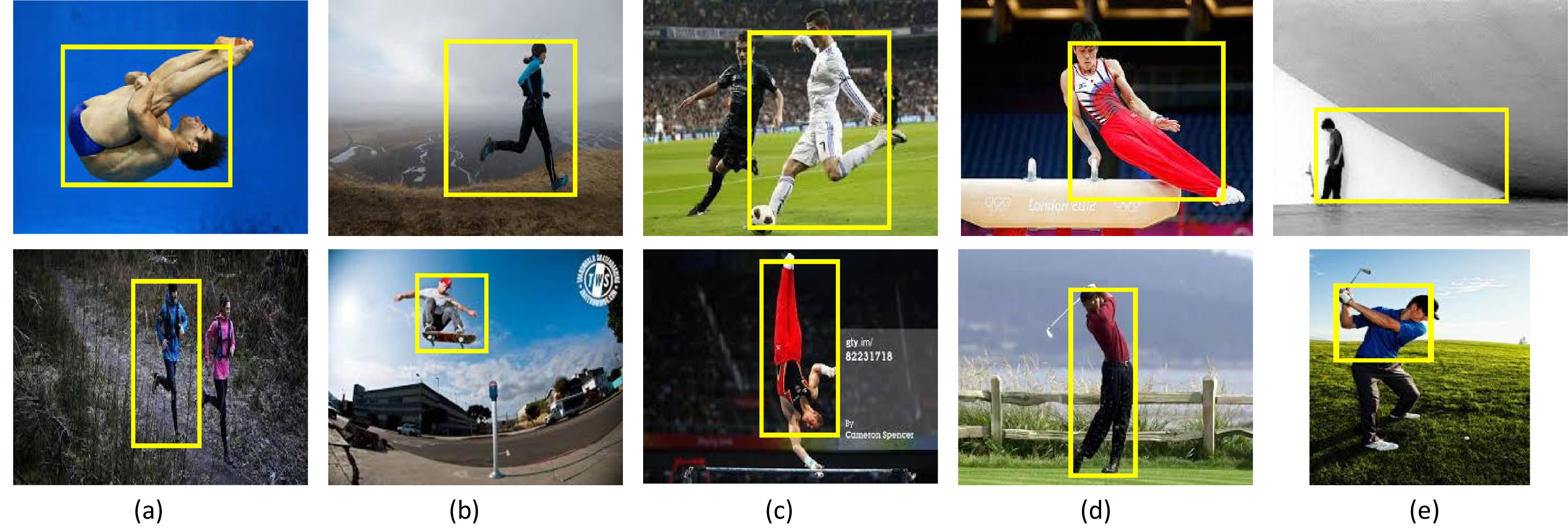}
\end{center}
   \caption{ Automatically obtained bounding boxes using multiple images. We used these automatically obtained bounding boxes to train our actioness detector. Last column (right) shows the failure cases.}
 \label{Coll_Image_Vis}
\end{figure*}

A few qualitative examples of unsupervised bounding box extraction for these images are shown in Fig.  \ref{Coll_Image_Vis}.

\paragraph{Negative Samples for Actioness Detection}

Given the video action proposals in each video, we find the ones that have very low optical flow derivatives and hence probably do not contain an action. We sample patches from those proposals and use them as negative data for training actionness detector.  In practice, we found these patches are more useful compared to non-action images from Google.

\paragraph{Image Actionness Detector}

Given the positive and negative samples( see Fig. \ref{actionness}), we fine-tune  Alexnet \cite{AlexNet} to obtain actionness detector. We use around 6,000 positive and negative samples for training and 3000 positive and negative samples for validation. For training, the batch size is 256, the learning rate is $10^{-3}$ and number of iterations is $50K$. The output of this classifier is the probability of each patch being representative of an action. Finally, the actionness score of sub-proposal is the mean of actionness score of each individual patch within sub-proposal.

\subsubsection{Motion Score}
\label{sec:motion_score}
Actionness  measure generic notion of actionnness within the bounding box of each sub-proposal but does not capture the explicit motion information. Due to large camera motion and background clutter, the straightforward use of optical flow magnitude within each proposal is noisy. Moreover, mostly, a large motion is generated by legs or hands and therefore, the proposals which capture these parts would get high motion scores. However, we want to assign high motion score to the proposals that enclose the complete moving body. Fortunately, motion edges produce high gradients across the whole moving body.  We build upon the observation from \cite{ZitnickECCV14edgeBoxes}, that higher the number of motion contours enclosed by proposal patch, the higher the probability the patch contains a complete moving body.

\begin{figure*}
\begin{center}
\includegraphics[width=10cm,height=3.5cm]{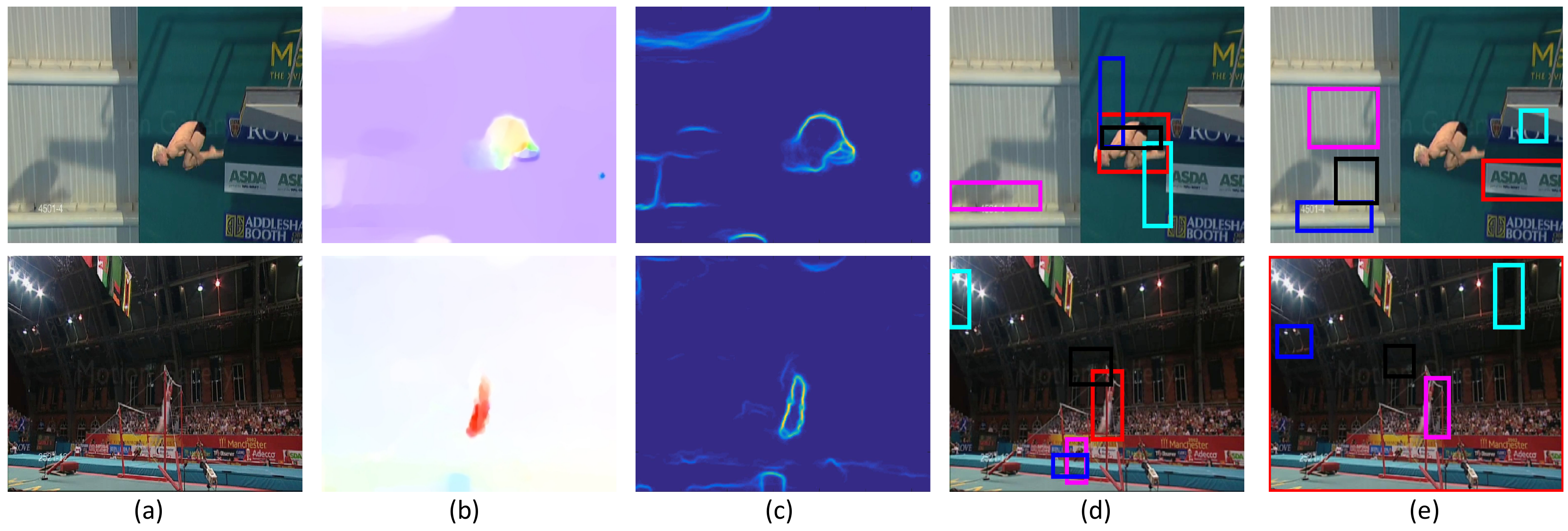}
\end{center}
   \caption{ (a) video frames, (b) optical flow, (c) motion edges, (d) proposals computed based on motion edges and (e) proposals computed based on image edges. The red bounding box shows the highest scored proposal.}
 \label{EdgeBox_Vis1}
\end{figure*}

Specifically, given the optical flow of two video frames (see Fig  \ref{EdgeBox_Vis1}.), we  compute motion edges for each channel of optical flow separately, The final motion edges are obtained as:  $E_m$=$E_u$ + $E_v$, where $E_u$ and $E_v$ represent motion edges of \textit{u} and \textit{v}  respectively. 
We, then, use an efficient framework of \cite{ZitnickECCV14edgeBoxes} to obtain several bounding boxes which represent the moving bodies in each frame.  Figure 4 (c) shows motion edges for shown video frames and Figure 4(d) shows bounding boxes obtained using these motion edges. In Figure 4(e), we show that the bounding boxes obtained using motion edges enclose the action better than the bounding boxes obtained using image edges.

To obtain motion score for any video proposal patch in a particular frame, we compute its overlap with all bounding boxes (obtained using motion edges) in that frame and assign it the score of the highest overlapped bounding box. Finally, the motion score of any sub-proposal is the mean of the motion score of its proposal patches.

\subsection{Edge Score}
\label{sec:edge_score}

The combined actionness and motion scores give the probability that a sub-proposal contains some action; however, it ignores the consistency and similarity between sub-proposal across different proposals. These similarities and consistencies are encoded in the edge score, which is a combination of sub-proposal appearance and shape similarities.

The shape similarity of two sub-proposal is measured by intersection-over-union (IOU) between last frame in one sub-proposal and the first frame of its temporally next sub-proposal. It is defined as:  
\begin{equation}
{\Psi^o = \frac{Area(b_{i,l} \cap b_{j,1})}{Area(b_{i,l} \cup b_{j,1})},}
\end{equation}
where $b_{i,l}$ represents bounding box in the last frame of a sub-proposal $i$ and $b_{j,1}$ represents the bounding box  in  the first frame of its temporally next sub-proposal $j$. To capture appearance similarity $\Psi^a$, we use Euclidean similarity between mean of HOGs of patches within each sub-proposal.


\section{Experiments}
\label{sec:exp}

The main goal of our experiments is to evaluate the accuracy of proposal ranking for trimmed and untrimmed datasets and validate proposal ranking effectiveness for action detection. In all the experiments, we used  $\lambda^i$=$\lambda^m$=$\lambda^o$=$\lambda^a$ =1 (in Equation (1) and (2)).

\subsection{Proposal Ranking in Trimmed Videos}
We evaluated the proposed approach on two challenging datasets: UCF-Sports \cite{action-mach-08} and sub-JHMDB \cite{Wang14}. 

\textbf{UCF-Sports} contains 10 human actions. This dataset includes actions such as diving, horse riding, running, walking, swinging,
etc. These videos contain large camera motion and dynamic backgrounds. 

\textbf{Sub-JHMDB} contains 316 video clips covering 12 human actions: climb, pullups, push, climb stairs, jump, catch, etc. This dataset is quite challenging due to clutter and variations in human motion and appearances.
\begin{figure*}[t]
\begin{center}
\includegraphics[width=13.5cm,height=7.2cm]{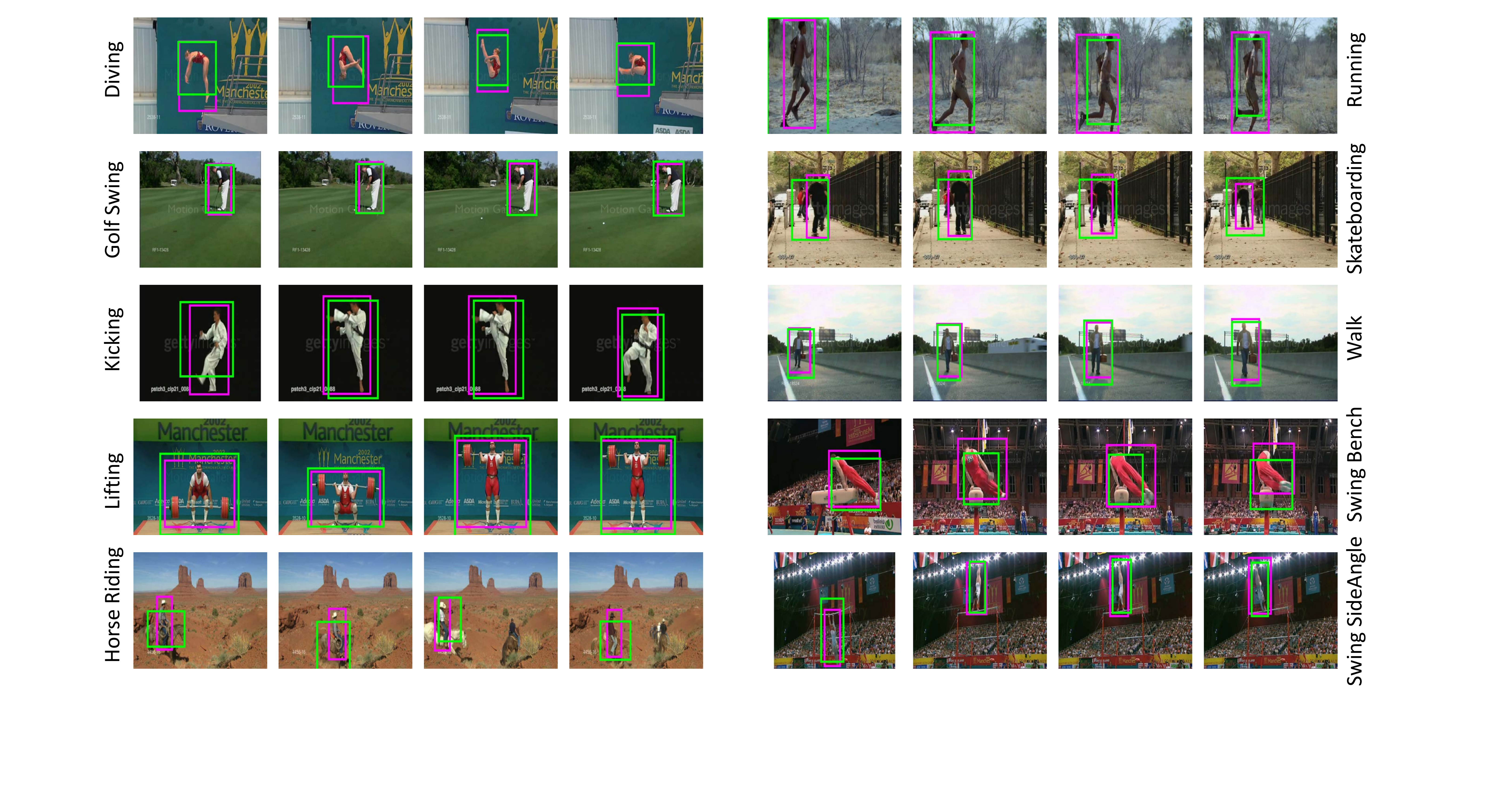}
\end{center}
   \caption{Qualitative results for UCF-Sports. Each row shows four frames of videos from two UCF-Sports actions. The magenta and green boxes respectively indicate ground truth and top ranked action proposal.}
 \vspace{-.1cm}
   \label{Exp_UCFSports}
\end{figure*}

 {In order to calculate localization accuracy, we compute overlap of top ranked action proposal with ground truth annotations as defined in \cite{Mihir}.}
 Since our approach is generic as it does not depend on any specific underlying action proposal method, we demonstrate localization results for three recent action proposals methods \cite{Danoneata,Mihir,APT}.  The method proposed in \cite{Danoneata,Mihir} are based on supervoxel segmentation and hierarchical merging of those segments. While van Gemert et al. \cite{APT} used improved dense trajectories \cite{IIDTF} clustering to obtain action proposals. We first compute optical flow derivatives for each proposal and remove highly overlapped proposals using non-maximal suppression.

 We compared our approach with two recently proposed \textit{weakly supervised} proposal ranking methods \cite{TangCVPR13,NegativeMining}. These methods achieve proposals ranking by exploiting proposals from negative videos, i.e., by using proposals from the videos that contain actions other than the action of interest.  Specifically, \cite{NegativeMining} ranked the proposals according to their nearest neighbor in negative videos. Larger the distance of proposal to the nearest neighbor proposal in the negative videos, the higher is the rank of the proposal. The method in  \cite{TangCVPR13} improved over \cite{NegativeMining} and employed all negatives proposals and penalize proposals to be ranked using their distance to negative proposals.  We implemented both methods as described in \cite{TangCVPR13}. For proposals representation, we used CNN features \cite{vedaldi15matconvnet} within proposal patch averaging over sample frames in the proposal, similar to  \cite{Mihir_15}.

\begin{figure*}
\begin{center}
\includegraphics[width=13.5cm,height=7.2cm]{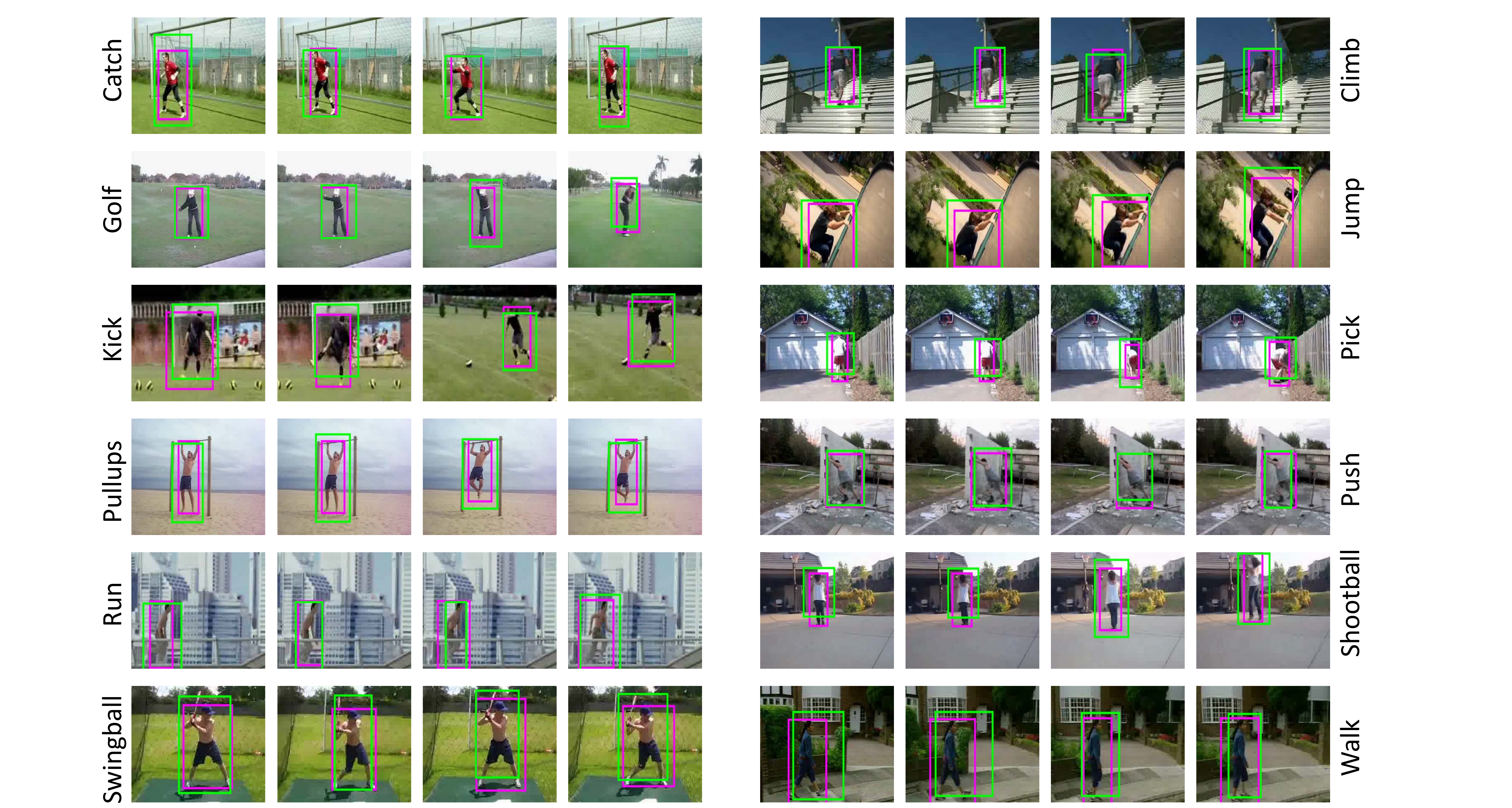}
\end{center}
   \caption{Qualitative examples from Sub-JHMDB. Each block shows four frames from a different action video in Sub-JHMDB. The magenta and green boxes respectively indicate ground truth and top ranked action proposals.}
 \label{Exp_JHMDB}
\end{figure*}

In Table  \ref{table_exp_1}, we demonstrate localization accuracy of top most proposal  over different thresholds. We compared our approach with both methods \cite{TangCVPR13,NegativeMining}  for all proposals methods we used.  We do not use Jain et al. \cite{Mihir} proposals for Sub-JHMDB, due unavailability of their code for this dataset. Since \cite{Danoneata,Mihir, APT} do not provide proposal ranking, we randomly sample proposal from each video and report their localization accuracy (averaged over three iterations).

\begin{figure*}
\begin{center}
\includegraphics[width=12cm,height=5cm]{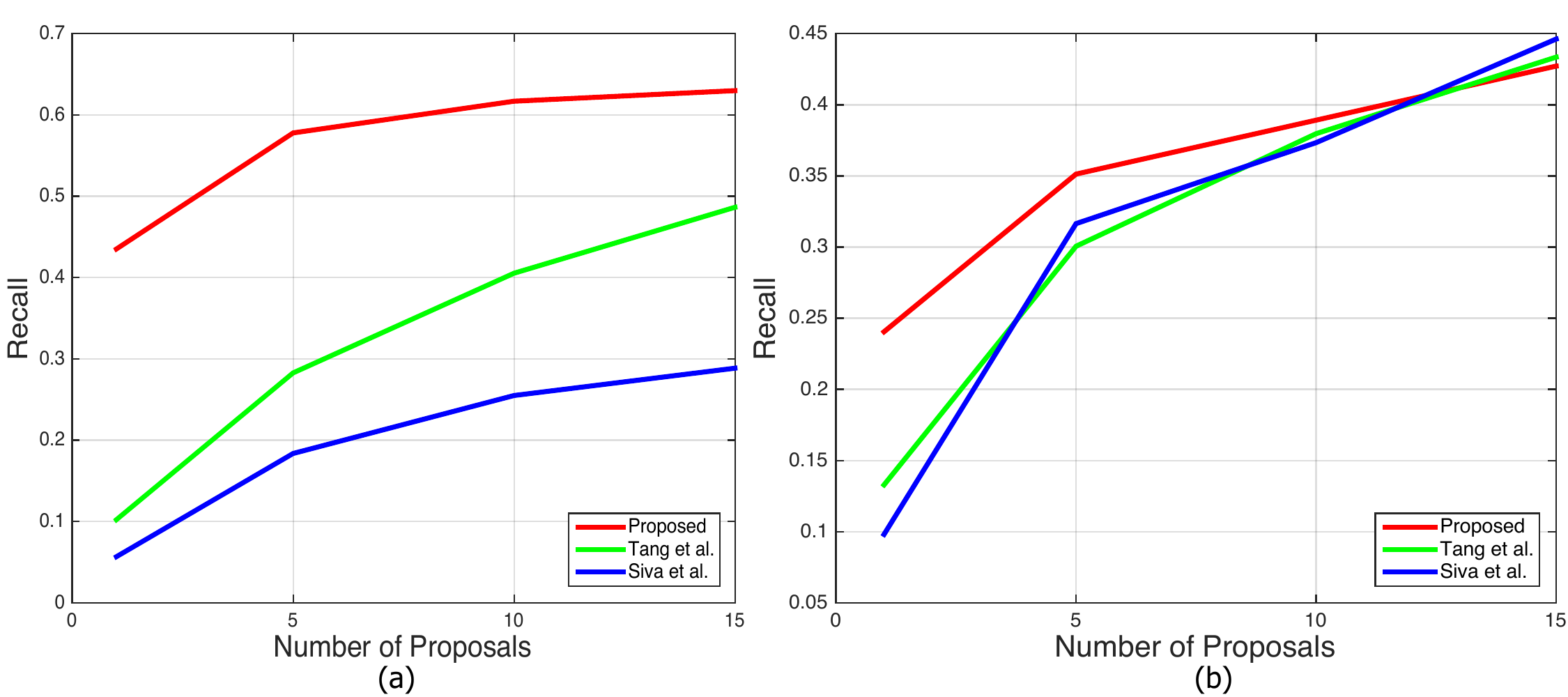}
\end{center}
   \caption{Recall versus number of proposals for UCF-Sports (a) and sub-JHMDB (b). The results are shown for Proposed method (red), Tang et. al. (green) and Siva et. al. (blue). }
   \label{Recall_NumofProposals}
\end{figure*}

It can be seen in Table \ref{table_exp_1} that for both datasets and for all proposal methods, our method significantly outperforms baselines without even using video level labels. As we increase the threshold, the localization accuracy decreases; however, our results are always better compared to the baselines. At overlap threshold of 20\%, as used by several recent works for action detection \cite{Yicong-cvpr2013,Khurram,Mihir}, our method has more than 85\% accuracy for both datasets. {In Figure \ref{Recall_NumofProposals}, we demonstrate recall versus different number of proposals. The results
demonstrate that our approach significantly outperforms baselines at all different
number of proposals.} 

\begin{table}[b]
\begin{center}
\caption{{Contribution from different components on UCF-sports  with van Germert et al. proposals.}}
\resizebox{10cm}{!}{
\begin{tabular}{l*{7}{c}}
 \hline
 Components & t=0.1 & t=0.2 & t=0.3 & t=0.4 & t=0.5 & t=0.6 & MABO \\
 \hline
 Motion Score + Shape + Appearance   & 94.2 & 88.3 & 74.0 & 52.6 & 31.2 & 11.4 & 40.4 \\
 Actionness + Shape + Appearance & 81.2 & 65.6 & 45.4 & 23.4  & 14.9 &  7.8  & 28.9 \\
Actionness + Motion Score + Appearance & 96.1   &  92.9    & 83.8  &   64.9  &   47.4  &   16.9  & 44.9 \\
Actionness + Motion Score + Shape &  96.1    &  92.7    & 83.8    & 62.9    & 39.6   &   16.9  & 44.9 \\

  Motion Score + Actionness + Shape + Appearance        &  94.8  &  92.9  & 83.7  &  64.3  & 43.5  &  18.2  &  45.5  \\
 \hline
\end{tabular}
}
\label{table_exp_components}
\end{center}
\end{table}
Fig. \ref{Exp_UCFSports} and Fig. \ref{Exp_JHMDB} show the qualitative results for top-ranked proposals for different videos of UCF-Sports and Sub-JHMDB. It can be seen that top ranked proposals cover the actor very well despite the presence of many noisy proposals due to clutter, large camera motion, and dynamic backgrounds.

Table \ref{table_exp_components}. shows the contribution of actioness score and motion score for final action localization.  { Top row shows that using Motion score and Shape + Appearance similarities (for graph construction), we achieve 40.4 (MABO). Second-row shows using Actionness and Shape + Appearance gives 28.9. The last row demonstrates that motion and actionness have complementary information and combination of both results in total MABO of 45.5. Finally, third and forth rows show localization accuracies using individual graph edge component.  The results demonstrate that each component of our method is essential to achieve final localization accuracy.}

It is interesting to observe that through proposal recombination, in many videos, we are able to discover proposals that are better than the best existing proposal in the initial proposal method. Given top 20 proposals, for UCF Sports and Sub-JHMDB, in 64 and 83 videos respectively, our newly generated proposals are better than all underlying proposals \cite{Danoneata}.

\begin{table}
\begin{center}
\caption{Comparisons of Proposal Ranking using Different Action Proposal}
\resizebox{12cm}{!}{
\begin{tabular}{l*{9}{c}}
 \hline
 Datasets & Proposal & Methods & t=0.1 & t=0.2 & t=0.3 & t=0.4 & t=0.5 & t=0.6 & MABO \\
 \hline
 \multirow{9}{2em}{UCF-Sports} & \multirow{3}{7em}{Jain et al.} & \textbf{Ours} & \textbf {93.5} & \textbf{90.3} & \textbf{78.6}  & \textbf{58.4}  & \textbf{39.6} & \textbf{24.7} & \textbf{43.9}  \\
                          & & Tang et.al. & 67.4  &  50.4 &   38.5   &  22.9  &   15.7 &   9.6  & 25.1 \\
                          & & Siva et.al. & 66.9   &  48.9 &   35.3 &    20.3  &   12.0    & 6.0 & 23.3 \\
                          & & {Jain et.al.} & {30.5} & {22.6} & {18.4}  & {12.5}  & {8.8} & {4.6} & {12.2} \\
                          \cline{2-10}
                          & \multirow{3}{7em}{Oneata et al.}   & \textbf{Ours} & \textbf{95.4} & \textbf{89.6} & \textbf{79.2} & \textbf{61.0} & \textbf{44.2} & \textbf{19.5} & \textbf{44.7} \\
                          & & Tang et.al. & 70.4    & 47.9 &   35.2   &  23.2   &  8.4    & 5.6  & 24.3 \\
                          & & Siva et.al. & 84.8   &  61.4  &  41.4   & 25.5    & 11.0   & 2.1 & 27.5 \\
                          & & {Oneata et al.} & {70.6} & {45.9} & {25.8} &  {14.7} &  {7.8} &  {3.5} & {21.2} \\
                          
                          \cline{2-10}
                          & \multirow{3}{7em}{van Gemert et al.}  & \textbf{Ours} &  \textbf{94.8} & \textbf{92.9} &\textbf{83.7} & \textbf{64.3} & \textbf{43.5} & \textbf{18.2} & \textbf{45.5} \\
                          & & Tang et.al. &69.3  &   47.4  &   34.3   &  19.7 &    10.2  &  4.4 &  23.6 \\
                          & & Siva et.al. & 80.8  &  51.1  &  31.2   &  17.7  &   5.7  &   1.4 & 23.3\\
                          & & { van Gemert et al.} &  {62.7}  &  {40.6}  & {22.3}  & {11.9} &    {6.7}   &   {3.7} & {19.0} \\
 \hline
 \multirow{6}{2em}{Sub-JHMDB} & \multirow{3}{7em}{Oneata et al.}   & \textbf{Ours} & \textbf{94.6}& \textbf{84.5} & \textbf{69.6} & \textbf{53.5} & \textbf{38.6} & \textbf{24.1} & \textbf{42.9} \\
                          & & Tang et.al. & 64.2  &  39.6  &  27.2 &   21.5 &   13.3   & 7.6 & 22.2 \\
                          & & Siva et.al. & 83.5   &    61.7   &    40.5    &  22.8        & 9.8   &   3.8 & 26.9 \\
                          & & {Oneata et al.} & {63.9}   & {36.7}   & {21.6} & {11.5} & {5.2} & {2.2}  & {18.2}\\
                          \cline{2-10}
                          & \multirow{3}{7em}{van Gemert et al.}  & \textbf{Ours} & \textbf{97.1}& \textbf{90.5} & \textbf{75.6} & \textbf{50.9} & \textbf{24.1} & \textbf{5.1} & \textbf{39.5} \\
                          & & Tang et.al. & 86.1  &  71.2   &  38.9   &  10.8  &  2.5  &  1.3  &    25.9 \\
                          & & Siva et.al. & 88.6  &   74.0   &  47.5  &   21.8   &  7.6  &   2.2  & 28.8\\
                          & & {van Gemert et al.} & {83.7}  & {58.1}  & {36.5} & {15.6} & {6.6}  & {1.9} & {25.1}\\
 \hline
\end{tabular}
}
\label{table_exp_1}
\end{center}
\end{table}


\subsection{Proposal Ranking in Untrimmed Videos}

In addition to trimmed datasets, we tested our approach on untrimmed \textbf{MSR-II} dataset. This dataset has 54 untrimmed videos and contains three actions: Handwaving, Clapping, and Boxing. This dataset is very challenging due to background clutter and occlusions. 
\begin{figure*}
\begin{center}
\includegraphics[width=8cm,height=6cm]{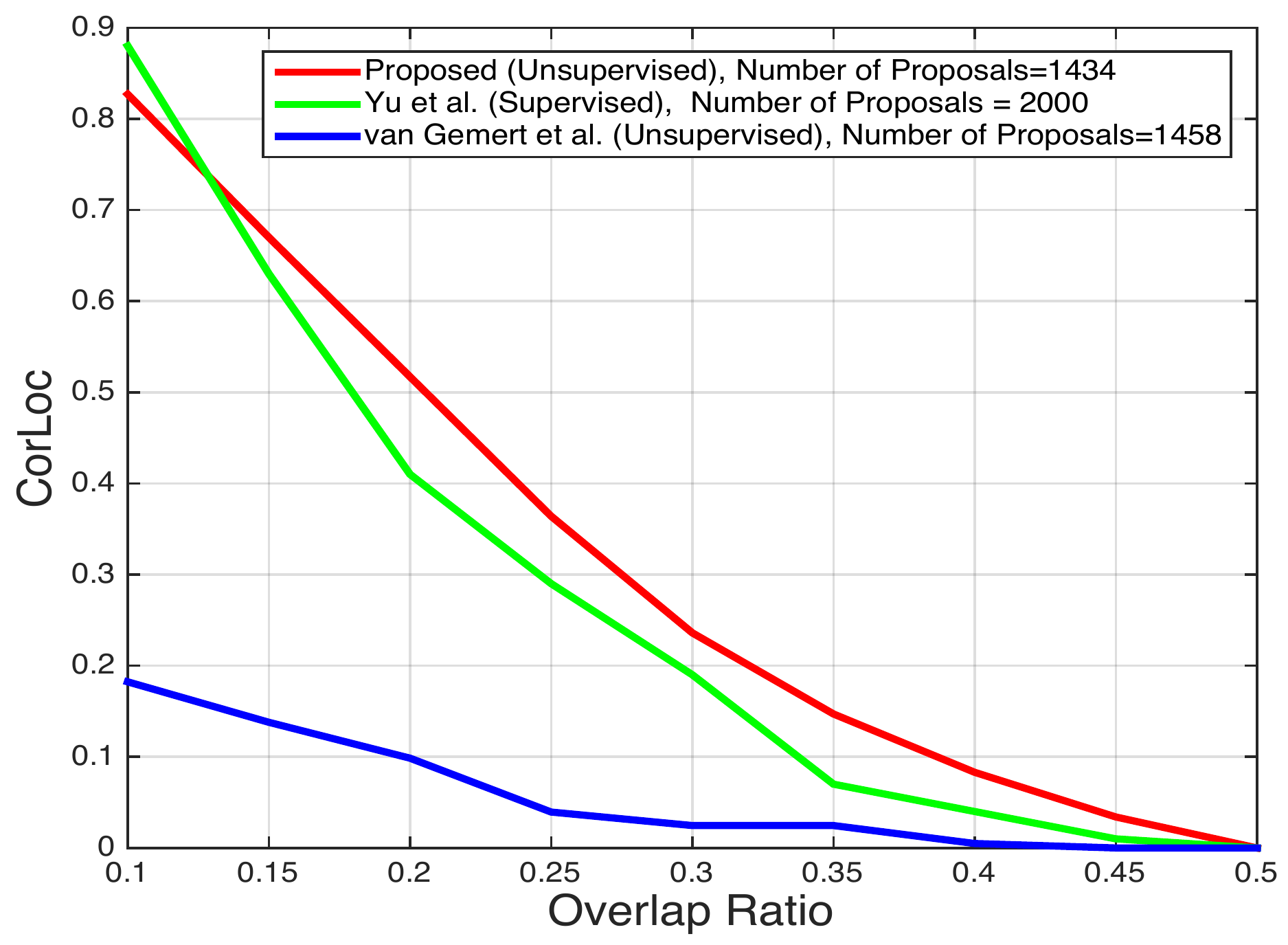}
\end{center}
   \caption{ {CorLoc comparision  for different thresholds on MSR-II.}}
   \label{MSR_CorLoc}
\end{figure*}

Since these videos are untrimmed, we first employ an unsupervised shot-detection method  \cite{ChangePoint} to divide the long video sequences to small shots and limit the action proposals within the shots. Below, we describe in details shot detection procedure.

{ \textbf{Temporal Segmentation: } We calculate the mean location of the action proposals in each video frame, and then the action proposal speed is calculated by the location distance between adjacent frames.  For action proposal diversity, we calculate the standard deviation of the locations of action proposals in every video frame. Finally, for the frame difference, we simply calculate the sum of the pixel differences between adjacent video frames. Given these three features, we employ a thoroughly studied
and well developed multiple change point detection scheme technique in statistic \cite{ChangePoint} for detecting shots.}  

{Let $\textbf {Z} \in \mathbb{R}^{ N \times K}$ represents the matrix obtained by horizontal  concatenation of features vectors of length K from N video frames. Note that in our case, since we have 3 features,  K=3. We seek to identify piece-wise approximation of  $\textbf {Z}$, i.e., $\textbf {X}$  using following convex optimization formulation:
\begin{equation}
   \displaystyle\min_{\substack{\bf X}} \frac{1}{2}\norm{\bf{Z}-\bf{X}}^2 +\lambda \sum_{n=1}^{N-1}\frac{ \norm{X_{n+1,.}-{X_{n,.}}}}{w_n},
  \end{equation} 
  where number of shots are automatically determined by convex total variation \cite{ChangePoint} and $w_n$ represents temporal location-dependent weights given as:
 \begin{equation}   
   w_n=\sqrt{\frac{N}{n(N-n)}}.
  \end{equation}
Note that the  first term in Equation. 1 represents quadratic error criterion and the second term enforce row-wise sparsity in X. We solve above equation efficiently using group LARS \cite{Modelselection}.}

On average, we have 9 shots (unsupervisedly selected)  in each video and we found top ranked proposals in each shot. We compared our method with recently proposed state-of-the-art supervised action proposals method \cite{FastProposals} and report CorLoc results in Figure.  \ref{MSR_CorLoc}.  It can be seen that even with 30\% less number of proposals, our \textit{unsupervised} approach performs better than the \textit{supervised} action proposal method. {Furthermore, we show the results of \cite{APT} (blue curve) with the number of proposals equal to our method (randomly selected). The quantitative results shown in Figure \ref{MSR_CorLoc} demonstrates that our method significantly outperform when considering less number of proposals.}

In the next section, we discuss the action detection using re-ranked proposals.

\subsection{Action Detection}

\begin{figure*}
\begin{center}
\includegraphics[width=12cm,height=5.5cm]{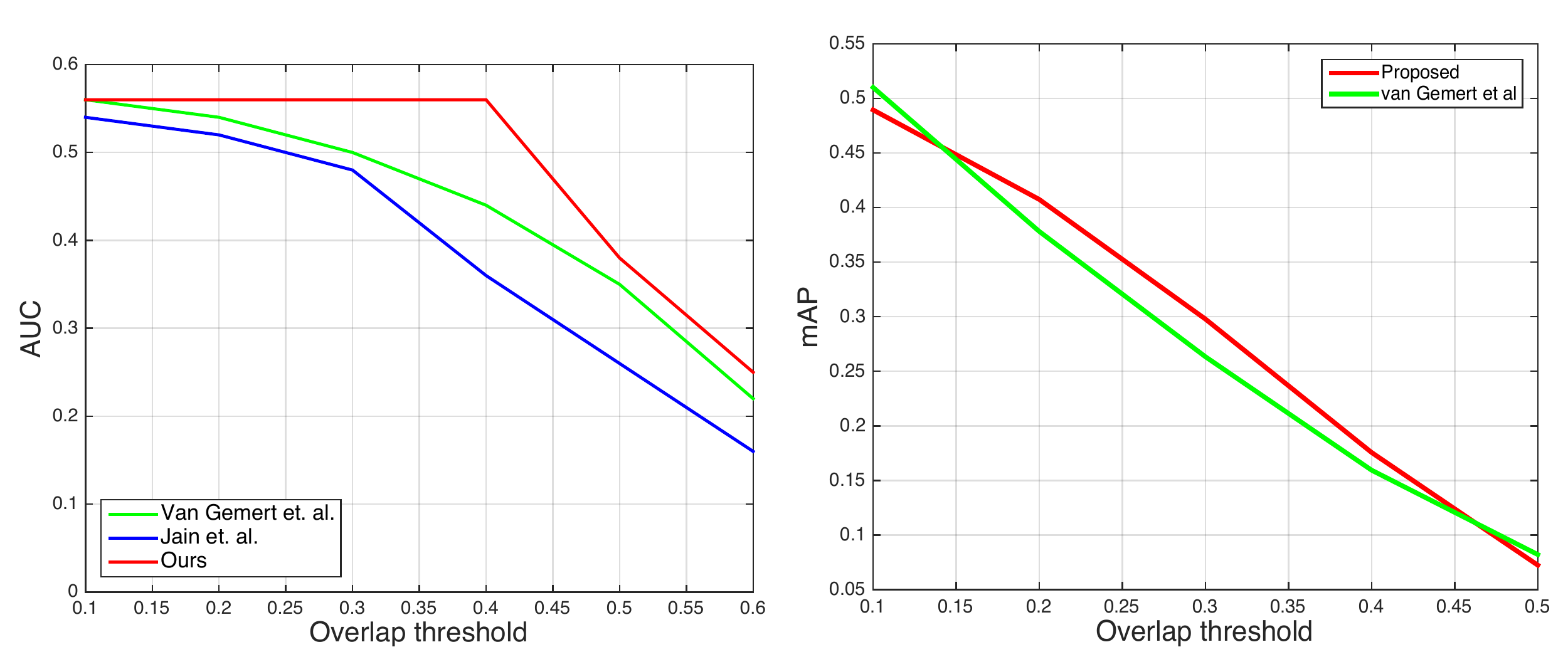}
\end{center}
   \caption{{Left: The AUC curves for UCF-Sports dataset \cite{action-mach-08}. The results are shown for proposed method (red), Jain et al. \cite{Mihir} (blue) and  Van Gemert et al. \cite{APT} (green). Right: mAP curves are shown for Proposed method (red), van  Van Gemert et al. \cite{APT} (green) }}
     \label{ROC_UCFSports_AL}
\end{figure*}

Better-ranked action proposals not only improve computation efficiency by reducing the search space but also improve action recognition accuracy by reducing the false positive rate.  We closely followed the procedure described in  \cite{APT} for proposal classification.  Specifically, we concatenate dense trajectory features (MBH, HOG, HOF, Traj) into 426-dimensional vector and reduce its dimension to half using PCA. We randomly sample features from training videos and fit 128 GMMs to it. We compute fisher vector representation of each proposal followed by power and  $L_2$ normalization. Finally, we used a linear classifier for training action detector. In our case, during testing, we only used top 25 proposals for features computation as well as classification. The final prediction scores of top 25 proposals are obtained by multiply proposal scores and classifier scores.  In Fig. \ref{ROC_UCFSports_AL}, we compared our results with two recent state-of-the-art proposal based action detection methods. It can be seen that we outperform both methods with significant margin at all overlapping thresholds. 

{In addition to UCF-Sports, we also demonstrate results for action detection results on UCF101. We use \cite{APT} for getting initial proposals. Experimental results in Figure \ref{ROC_UCFSports_AL} shows the superiority of our approach as compared to baseline \cite{APT}. Note that author in  \cite{APT} have used 2299 proposals in each video. In  contrast, our method just use only 200 proposals in each video.}

Performing better than the baselines emphasizes the strength of the proposed approach and reinforces that properly ranked action proposals have significant impact on action detection.
\subsection{Computation Time:}

Our approach works on top of [7][4][5]. These proposals  take few minutes for computing proposal in each video.  We now have optimized our code slightly. Our code takes 0.002 seconds to compute action score for each proposal patch and 0.02 seconds for optimal path discovery for both UCF-Sports and sub-JHMDB datasets. Other steps: optical flow derivatives, HOGs similarity also take less than a sec.

\section{Conclusions}
\label{sec:conclusions}

We have proposed a method for action proposal ranking which aims to generate fewer but better action proposals which are ranked properly. We have tested our method on UCF-Sports, sub-JHMDB and MSR-II for action proposal ranking and compared it to several baseline methods. We have shown that the proposed method generates fewer but better proposals. The method is useful for many applications and can reduce the search space dramatically. We also performed experiments for action detection and showed better performance compared with state-of-the-art proposals based action detection methods.






\end{document}